\documentclass{ifacconf}

\usepackage{graphicx}      
\usepackage{natbib}        
\usepackage{amsmath}
\usepackage{amssymb}
\usepackage{comment}
\usepackage{bm}
\usepackage{float}
\usepackage{booktabs} 
\usepackage{booktabs} 
\usepackage{arydshln} 
\usepackage{todonotes}
\usepackage{soul}

\newcommand{\Yab}{\mathbf{Y^{*}}}
\newcommand{\ya}{y^{*}}

\newcommand{\ytb}{\mathbf{\Tilde{y}^{*}}}

\newcommand{\Cc}{\mathcal{C}}

\begin{document}
\begin{frontmatter}

\title{Training Neural ODEs Using Fully Discretized Simultaneous Optimization\thanksref{footnoteinfo}} 

\thanks[footnoteinfo]{Support from a BASF/Royal Academy of Engineering Senior Research Fellowship is gratefully acknowledged. }

\author[First]{Mariia Shapovalova} 
\author[First]{Calvin Tsay} 

\address[First]{Department of Computing, Imperial College London, London, UK (e-mail: c.tsay@imperial.ac.uk).}

\begin{abstract}
Neural Ordinary Differential Equations (Neural ODEs) represent continuous-time dynamics with neural networks, offering advancements for modeling and control tasks.
However, training Neural ODEs requires solving differential equations at each epoch, leading to high computational costs.
This work investigates simultaneous optimization methods as a faster training alternative. 
In particular, we employ a collocation-based, fully discretized formulation and use IPOPT---a solver for large-scale nonlinear optimization---to simultaneously optimize collocation coefficients and neural network parameters. 
Using the Van der Pol Oscillator as a case study, we demonstrate faster convergence compared to traditional training methods. 
Furthermore, we introduce a decomposition framework utilizing Alternating Direction Method of Multipliers (ADMM) to effectively coordinate sub-models among data batches. 
Our results show significant potential for (collocation-based) simultaneous Neural ODE training pipelines.
\end{abstract}


\end{frontmatter}

\section{Introduction}

Data-driven dynamic models are increasingly prevalent in chemical and process systems engineering, providing useful alternatives to first-principles, physical models~\citep{bhosekar2018advances,thebelt2022maximizing}. 
In particular, neural network-based models are popular surrogates in scheduling and control applications owing to their flexibility and representation power, e.g., as scale-bridging models~\citep{tsay2019110th}. 
Neural networks can have various architectures and consist of learnable weights and biases that are optimized during training by minimizing a specified loss function. 
This is commonly achieved using gradient-based algorithms, which iteratively update the model parameters by taking steps in the negative direction of the gradient of the loss function~\citep{deep_learning}.

Neural Ordinary Differential Equations (Neural ODEs) ~\citep{Chen:2018} bridge neural networks with dynamical systems modeling, leveraging existing knowledge of ODEs. 
These models extend traditional networks to model unknown continuous-time dynamics by parameterizing the evolution of system states as a differential equation:
\begin{equation}\label{eq:neural_ode}
\frac{d\bm{x}(t)}{dt} = f_{\bm{\theta}}(\bm{x}(t), t), 
\end{equation}
where $\bm{x}(t)$ represents the state at time $t$, and $f_{\bm{\theta}}$ is the neural network parameterized by $\bm{\theta}$. Compared to more standard recurrent or convolutional neural networks, neural ODEs are flexible and can incorporate arbitrary time spacings. 
The framework has found numerous applications, including process control~\citep{luo2023model}, reaction modeling~\citep{sorourifar2023physics}, and parameter estimation~\citep{bradley2021two,dua2012simultaneous}.\\

Despite their advantages, training Neural ODEs (or making predictions) requires solving an initial value problem (IVP) at each iteration using a numerical ODE solver. Given the initial condition $\bm{x}(t_0)$, the state at time $T \geq t_0$ is computed by solving the differential equation: 
\begin{equation} \label{eq:odesolve}
\begin{aligned}
\bm{x}(T) &= \bm{x}(t_0) + \int_{t_0}^{T} f_{\bm{\theta}}(\bm{x}(t), t) \, dt \\
&= \text{ODESolve}(\bm{x}(t_0), f_{\bm{\theta}}, t_0, T),
\end{aligned}
\end{equation}
where ODESolve is a numerical IVP solver, and $t_0$, $T$ are the beginning and end of the integration interval, respectively. 
Training is typically based on the accuracy of the predictions $\bm{x}(T)$, requiring the numerical solution of \eqref{eq:odesolve} and backpropagation of gradients through the IVP solver at every iteration. 
These requirements lead to long training times of Neural ODEs ~\citep{Lehtim:2024:accelerated}. 

Given the above, this work uses \textit{spectral} numerical methods, specifically collocation, for the time integration of differential equations in Neural ODE training. 
Spectral methods offer several advantages: they are global as they approximate over the entire domain, display exponential convergence for smooth problems, and have better accuracy with a small number of points ~\citep{Boyd2000}. 
Spectral methods remain less explored than sequential methods in the context of Neural ODEs and have to date mostly been limited to approximating derivative targets ~\citep{Roesch2021Collocation}, or for non-simultaneous training ~\citep{Quaglino:2020:SNODE}. 
The novelty of this paper is that we show collocation can be employed in a \textit{simultaneous} optimization approach, i.e., the system dynamics are solved as equality constraints rather than by iterative simulation, for fast and stable Neural ODE training. 
Furthermore, we show that the proposed method may produce more parsimonious models and is amenable to batching via ADMM.

\section{Neural ODEs for Time Series}

Neural ODEs can be applied in various contexts, e.g., in generative modeling or as implicit layers in larger models. 
We focus on the basic, control-relevant setting of modeling time-series data comprising observations $\{\mathbf{y}_i, t_i\}_{i=0}^{N - 1}$, where $\mathbf{y}_i \in \mathbb{R}^d$ is the data vector at time $t_i$, and $T = t_{N-1}$ is the end of the time (and integration) interval. 
Our objective is to learn a parametric ODE model that, when integrated from an initial condition, results in a continuous trajectory $\mathbf{y}(t)$ approximating the observed data:
\begin{equation}\label{eq:NODE}
    \mathbf{y}'(t) = f_{\bm{\theta}}(\mathbf{y}(t), t), \quad \mathbf{y}(t_0) = \mathbf{y}_0, \quad t \geq t_0,
\end{equation}
where $\mathbf{y}'(t)$ is the time derivative of the system state $\mathbf{y}(t)$ and $f_{\boldsymbol{\theta}}$ is a neural network parameterized by $\boldsymbol{\theta}$. 
A trained Neural ODE model may also predict beyond the observed data interval $[t_0, T]$, provided the ODE solution remains valid and $f_{\boldsymbol{\theta}}$ satisfies conditions such as Lipschitz continuity.

During training, the solution $\mathbf{\hat{Y}}$ is computed by numerically solving the IVP:
\begin{equation}\label{eq:ODE_solve}
\mathbf{\hat{Y}} = \text{ODESolve}\left( f_{\bm{\theta}}, \mathbf{y}_0, \mathbf{t} \right) \in \mathbb{R}^{N \times d},
\end{equation}
where $\mathbf{t} = \{ t_i \}_{i=0}^{N-1}$ is the vector of time points matching the observations.
The model parameters, $\bm{\theta}$, are learned by minimizing a loss function that captures the discrepancy between the predictions and observed data. The mean squared error (MSE) loss function is a common choice for continuous-output regression:
\begin{equation}\label{eq:loss_func}
    \mathcal{L}_{\bm{\theta}}(\hat{\mathbf{Y}}, \mathbf{Y}) = \frac{1}{N} \left\Vert \mathbf{\hat{Y}} - \mathbf{Y} \right\Vert_F^2,
\end{equation}
where
$\mathbf{Y} \in \mathbb{R}^{N \times d}$ is the matrix of observed values and
$\| \cdot \|_F$ is the Frobenius norm. 
In summary, the goal of Neural ODE training can be formulated as computing $\bm{\theta}$ such that \eqref{eq:NODE} holds while minimizing \eqref{eq:loss_func}. 

\subsection{Sequential ODE Solvers}
In a typical Neural ODE training pipeline, we ensure \eqref{eq:NODE} holds by solving the ODE system in every iteration. In this \textit{sequential} approach, the solver computes $\mathbf{\hat{Y}}$ iteratively, e.g., by time stepping, until the end of the interval is reached at $T$. 
The simplest example is Euler's method, while more commonly used schemes are the Runge-Kutta methods, used as default solvers in \texttt{torchdiffeq} within PyTorch and \texttt{Diffrax} within JAX. We can generalize a step of a sequential numerical scheme as:
\begin{equation*}
    \mathbf{y}(t + h) = \mathbf{y}(t) + h \cdot \Phi(f, \mathbf{y}(t), t, h),
\end{equation*}
where $h$ is the step size, $f$ is the derivative function, and $\Phi$ denotes the (typically explicit) function that approximates the change in $\mathbf{y}$ over the interval from $t$ to $t + h$.

While this framework enables using tailored simulation methods for \eqref{eq:NODE}, using sequential ODE solvers for training poses several challenges. 
First, numerical errors can accumulate at each integration step, resulting in substantial global errors. 
Although adaptive solvers help control these errors, they add computational overhead and may still behave unpredictably on unseen data. 
Second, in addition to simulation CPU times, storing intermediate solutions for backpropagation requires significant memory. 
Despite the adjoint method ~\citep{Chen:2018} ensuring constant memory cost, it can substantially prolong training time.

\section{Spectral Methods}

As an alternative to the above, spectral numerical methods approximate the ODE solution as a linear combination of basis functions, e.g., trigonometric or orthogonal polynomials. The function coefficients are fitted over the integration domain, offering high accuracy and convergence rates for smooth problems~\citep{Boyd2000}.

\subsection{Collocation with Lagrange Interpolation}

 Collocation is a class of spectral methods in which an ODE is enforced at a set of discrete points, termed the \textit{collocation grid}, which we introduce as $\bm{\xi} = \{\xi_i\}_{i=0}^{N-1}$ in $[t_0, T]$. Under this framework, the approximate solution is
\begin{equation*}
    \mathbf{\Tilde{y}}(t) = \sum_{i=0}^{N-1} \bm{\beta}_i \phi_i(t),
\end{equation*}
where $ \{ \phi_i(t) \}_{i=0}^{N-1}$ represent the basis functions, $ \{\bm{\beta}_i\}_{i=0}^{N - 1} $
 are the coefficients to be determined, and $ t \in [t_0, T]$. 
 
We employ the barycentric form of Lagrange polynomials as basis functions due to its numerical stability and adaptability to diverse functions, including non-periodic behaviors ~\citep{Berrut2004BarycentricLI}. 
The use of Lagrange polynomials also offers an implementation simplification due to the interpolation property, which ensures that the coefficients coincide with the true state values at the collocation grid, such that $ \bm{\beta}_i = \mathbf{y}_i$ at each collocation point.
As a result, the approximated solution becomes:
\begin{equation}\label{eq:lagrange_interpolation}
\mathbf{\Tilde{y}}(t) = \sum_{i=0}^{N-1} \mathbf{y}_i \ell_i(t),
\end{equation}
where $\{ \ell_i(t) \}_{i=0}^{N-1}$ are the Lagrange basis functions and $\{ \mathbf{y}_i \}_{i=0}^{N-1}$ are the unknown coefficients, which are also the state values at each point $\xi_i$. 
We assume $\mathbf{y}$ to be unknown, given the presence of noise in real-life systems. 
By treating all $\mathbf{y}_i$ as coefficients~\citep{Berrut2004BarycentricLI} and differentiating the interpolation formula \eqref{eq:lagrange_interpolation}, we obtain:
\begin{equation}\label{eq:lagrange_derivative} 
\tilde{\mathbf{y}}'(t) = \sum_{i=0}^{N-1} \mathbf{y}_i \ell_i'(t). \end{equation}

Substituting \eqref{eq:lagrange_interpolation} and \eqref{eq:lagrange_derivative} into the Neural ODE \eqref{eq:NODE} and evaluating at each collocation point $\xi_i$ yields:
\begin{equation} 
\sum_{j=0}^{N-1} \mathbf{y}_j \ell_j'(\xi_i) = f_{\bm{\theta}}\left(\sum_{j=0}^{N-1} \mathbf{y}_j \ell_j(\xi_i), \xi_i\right), \quad i = 0, \ldots, N - 1. 
\end{equation}

This results in a system of nonlinear equations with respect to the unknowns $\{{\mathbf{y}_j}\}_{j=0}^{N-1}$. 
The system can be expressed in matrix form:
\begin{equation}\label{eq:col_alg}
\mathbf{D Y} = \mathbf{F}_{\bm{\theta}}(\mathbf{Y}, \bm{\xi}), 
\end{equation}

where: 
\begin{itemize}
    \item $\mathbf{D} \in \mathbb{R}^{N \times N}$ is the differentiation matrix with elements $D_{ij} = \ell_j'(\xi_i)$.
    \item $\mathbf{Y} = [\mathbf{y}_0, \dots, \mathbf{y}_{N-1}]^\top \in \mathbb{R}^{N \times d}$ is the matrix of unknown coefficients (true state values).
    \item $\mathbf{F}_{\bm{\theta}}(\mathbf{Y}, \bm{\xi}) = [f_{\theta}(\mathbf{\Tilde{y}}(\xi_0), \xi_0), \dots, f_{\theta}(\mathbf{\Tilde{y}}(\xi_{N-1}), \xi_{N-1})]^\top \in \mathbb{R}^{N \times d}$ contains the Neural ODE evaluated at each collocation point. 
\end{itemize}

Using the barycentric formula~\citep{Berrut2004BarycentricLI}, the differentiation matrix is defined as:
\begin{equation*}
D_{ij} = \begin{cases}
\displaystyle \frac{w_j}{w_i} \frac{1}{\xi_i - \xi_j}, & \text{if } i \ne j, \\
\displaystyle -\sum_{\substack{k=0 \\ k \ne i}}^{N-1} D_{ik}, & \text{if } i = j,
\end{cases}
\end{equation*}
where the weight is computed as:
\begin{equation*} 
w_i = \frac{1}{\displaystyle \prod_{\substack{k=0 \ k \ne i}}^{N-1} (\xi_i - \xi_k)}. 
\end{equation*}

The selection of the collocation grid significantly impacts the accuracy of the method. 
To mitigate errors caused by Runge's phenomenon, Chebyshev nodes of the second kind in $[-1, 1]$ are often used:
\begin{equation} \xi_i = \cos\left(\frac{i \pi}{N - 1}\right), \quad i = 0, \dots, N -1. \end{equation}

We refer the interested reader to \cite{YOUNG20191033} for a comprehensive discussion of collocation grids.

\section{Proposed Methodology}

So far, we have converted the continuous ODE problem into a discrete algebraic system \eqref{eq:col_alg} by incorporating collocation and Lagrange interpolation. 
Our goal is to minimize the loss function \eqref{eq:loss_func} while enforcing that the collocation-estimated derivatives, computed as $\mathbf{D Y}$, match the neural-network-predicted derivatives $\mathbf{F}_{\bm{\theta}}(\mathbf{Y}, \bm{\xi})$ at the collocation points. 
The challenge arises because both the true values of the state values $\mathbf{Y}$ and the parameters $\bm{\theta}$ of the neural network are unknown. 

\subsection{Simultaneous Approach}
Our proposed approach is to incorporate the collocation system \eqref{eq:loss_func} as equality constraints in a single nonlinear optimization framework, where the objective function captures the discrepancy between the observed and optimized states, e.g., the MSE.
By solving for $\mathbf{Y}$ and $\bm{\theta}$ simultaneously, we effectively train the neural network from observed data while enforcing the collocation constraints.

The \textit{simultaneous} approach in the context of collocation-based dynamic optimization can be further explored in the works by \cite{param_estimation_ipopt, KAMESWARAN20061560}, where it is applied to the problem of parameter-estimation in differential equation systems.

\subsection{Implementation}
To implement this methodology, we utilize the Interior Point OPTimizer (IPOPT), which is well-suited for solving continuous, large-scale nonlinear optimization problems~\citep{BIEGLER2009575}. 
For the software implementation, we call IPOPT through the open-source \textit{Pyomo} algebraic modeling language.
Recent research ~\citep{OMLT} demonstrates how neural networks can be represented as constraints within the Pyomo framework.\\

We initialize two optimization variable groups within Pyomo: \textit{state variables} and \textit{neural network parameters}.

\begin{itemize}
    \item \textbf{State Variables}, $\Yab$, represent the system's state at collocation time points. 
    To expedite training, the state variables can also be initialized using smoothed observed data. These variables aim to approximate the true values $\mathbf{Y}$ from the observed values $\mathbf{Y}_{\text{obs}}$.
    \item \textbf{Neural Network Parameters}, $ \bm{\theta} $,  include the weights and biases of the neural network. 
\end{itemize}

The objective function captures the difference between the observed data and the state variables approximated by the collocation equations, instead of the output of ODESolve as in \eqref{eq:ODE_solve}. 
We express the objective function as a combination of the MSE loss and regularization terms:
\begin{equation}\label{eq:proposed_loss}
\mathcal{L}(\Yab, \mathbf{Y}_{obs}) = \frac{1}{N} \|\Yab - \mathbf{Y}_{obs}\|_F^2 + \lambda \|\bm{\theta}\|_2^2,
\end{equation}

\noindent where 
\begin{itemize}
    \item $\Yab \in \mathbb{R}^{N \times d}$ is the matrix of estimated state variables (the variables being optimized).
    \item $\mathbf{Y}_{obs} \in \mathbb{R}^{N \times d}$ is the matrix of observed values.
    \item $\bm{\theta}$ is the vector of neural network parameters.
    \item $\|\cdot\|_2$ and $\|\cdot\|_F$ are the Euclidean (L2) and Frobenius norms respectively.
    \item $\lambda$ is the regularization parameter.
\end{itemize}

We enforce consistency between the neural network and the derivative of $\Yab$ at each collocation point $\xi_i$:
\begin{equation*}\label{eq:proposed_constraints}
\sum_{j=0}^{N-1} \mathbf{y}^{*}_j l_j'(\xi_i) = f_{\bm{\theta}}(\mathbf{y}^{*}_i, \xi_i).
\end{equation*}
Here, $\ell_j'(\xi_i)$ is the element $D_{ij}$ of the differentiation matrix, so the left-hand side approximates the derivative of the optimized state values. The right-hand side is the output of the neural network. 

\subsection{Problem Formulation}
We formulate the optimization problem as follows:
\begin{equation*}
\begin{aligned}
& \min_{\Yab, \bm{\theta}} \mathcal{L}(\Yab, \mathbf{Y}_{obs}), \\
& \text{Subject to:} \\
& \quad \text{Equality Constraints: }\mathbf{D \Yab} = \mathbf{F}_{\bm{\theta}}(\Yab, \bm{\xi}),\\
& \quad \text{Bounds: } \ya_L \leq \ytb \leq \ya_U, \quad \theta_{L} \leq \bm{\theta} \leq \theta_{R} \\
\end{aligned}
\end{equation*}
where: 
\begin{itemize}
  \item $\mathcal{L}: $ is the loss function as described in \eqref{eq:proposed_loss}.
  \item $\Cc: \mathbf{D \Yab} = \mathbf{F}_{\bm{\theta}}(\Yab, \bm{\xi})$ is the matrix of equality constraints. 
  \item $\Yab \text{ and } \bm{\theta}$ represent decision variables.
  \item $\ya_L, \ya_U$ and $\theta_{L},  \theta_{U}$ are the respective values for their lower and upper bounds\\
\end{itemize}

After the model is solved to optimality, the neural network can be used as the RHS of an ODE in a sequential or collocation-based solver in the post-training context. 

\subsection{Alternating Direction Method of Multipliers (ADMM)}

One potential disadvantage of the above simultaneous framework is that the entire dataset must be handled in a single optimization problem, while many training pipelines divide data into batches to alleviate computational or memory burden. 
We propose using the Alternating Direction Method of Multipliers (ADMM) to enable multi-batching by coordinating the training of separate submodels. For two `batches,' the problem can be written as:
\begin{equation*}
\begin{aligned}
\label{eq:decomposition}
\min_{\bm{\theta}_1, \bm{\theta}_2} \quad &\mathcal{L}_1(\bm{\theta}_1, \mathbf{Y}_1) + \mathcal{L}_2(\bm{\theta}_2, \mathbf{Y}_2) \\
\mathrm{s.t.} \quad &\bm{\theta_1} = \bm{\bm{\theta_2}},
\end{aligned}
\end{equation*}
where $\mathbf{Y}_1$, $\mathbf{Y}_2$ are the batches of data, $\bm{\theta}_1$, $\bm{\theta}_2$ are vectors containing parameters of each sub-model, and $\mathcal{L}_1$, $\mathcal{L}_2$ are the loss functions. Notice the `linking' constraints $\bm{\theta_1} = \bm{\bm{\theta_2}}$ enforce a consensus model between the two data batches. 

ADMM decomposes the above problem without the linking constraints by updating the optimization parameters and a dual variable (Lagrange multiplier) in an iterative manner ~\citep{Boyd2010}. 
Without the constraints, the problem is effectively decomposed into independent subproblems $\mathrm{min}_{\theta_i}\ \mathcal{L}_i(\bm{\theta}_i, \mathbf{Y}_i)$. 
The loss functions for the subproblems are reformulated as follows: 
\begin{equation*}
\mathcal{L}_{\text{ADMM}, i} = \mathcal{L}_i(\bm{\theta}_i, \mathbf{X}_i) + \frac{\rho}{2} \left\| \bm{\theta}_i - \bm{\bar{\theta}}^{(k)} + \frac{\mathbf{u}_i}{\rho} \right\|_2^2, \quad \text{for } i = 1, 2,
\end{equation*}
where 
$\bm{\theta}_i$ are parameters of submodel $i$,
$\bm{\bar{\theta}}^{(k)}$ are the consensus parameters at the $k$-th ADMM iteration,  $\rho$ is a scalar penalty strength, and $\mathbf{u}_i$ are the dual variables associated with subproblem $i$. 
For two submodels, the consensus weights in the $k$-th iteration are given by:
\begin{equation*}
\bm{\bar{\theta}}^{(k)} = \frac{\bm{\theta}_1^{(k)} + \bm{\theta}_2^{(k)}}{2}
\end{equation*}

The dual variables for each submodel $i$ are updated in each iteration using:
\begin{equation*}
\mathbf{u}_i^{(k+1)} = \mathbf{u}_i^{(k)} + \rho (\bm{\theta_i}^{(k)} - \bm{\bar{\theta}}^{(k)}).
\end{equation*}

The model can be run for a fixed number of iterations or until convergence, which is defined by the magnitude of the primal residual, which we compute as: 
\begin{equation*} r_{\text{primal}}^{(k)} = \sum_{i=1}^{2} \| \bm{\theta}_i^{(k)} - \bar{\bm{\theta}}^{(k)} \|_2. \end{equation*}

This ADMM framework allows us to train larger models by decomposing the problem into smaller subproblems, but can also improve the model generalization by learning from multiple trajectories simultaneously.

\section{Experimental Results}
We compare the proposed collocation-based approach for training Neural ODEs to two benchmark sequential implementations: JAX (Diffrax) and PyTorch (torchdiffeq). Before training, we apply the following preprocessing steps:
\begin{itemize}
\item{\textbf{Spacing}:} We interpolate the training data to the chosen collocation grid, ensuring the data align with collocation points. The collocation grids are scaled according to the time range of the observed data. 
\item{\textbf{Noise}:} We simulate measurement noise by adding zero-centered Gaussian noise with $\sigma = 0.1$.
\item{\textbf{Initialization}:}  We utilize \textit{Xavier} initialization for the neural network weights ~\citep{glorot2010understanding}. 
For the simultaneous approach, the state variables $\Yab$ are initialized to the values of locally weighted polynomial regression ~\citep{Cleveland01091988}. 
\end{itemize}

\subsection{Case Study: Van der Pol Oscillator}
The forced Van der Pol Oscillator is a 2-D ODE system that can be represented as two coupled first-order equations:
\[
\begin{cases}
u' = v, \quad u_0 = 0\\
v' = \mu (1 - u^2) v - u + A\cos(\omega t), \quad v_0 = 1,
\end{cases}
\]

where $u$ is the displacement, $v$ is the velocity, $\omega$ is the angular frequency, $\mu$ is the damping parameter, and $A$ is the external periodic force. For our experiments, we set the initial conditions as $u_0 = 0$ and $v_0 = 1$. The remaining parameters are chosen as $\mu = 1$, $A = 1$ and $\omega = 1$.

\subsubsection{Training and Inference Procedure}
\begin{figure}[H]
    \centering
    \includegraphics[width=1\linewidth]{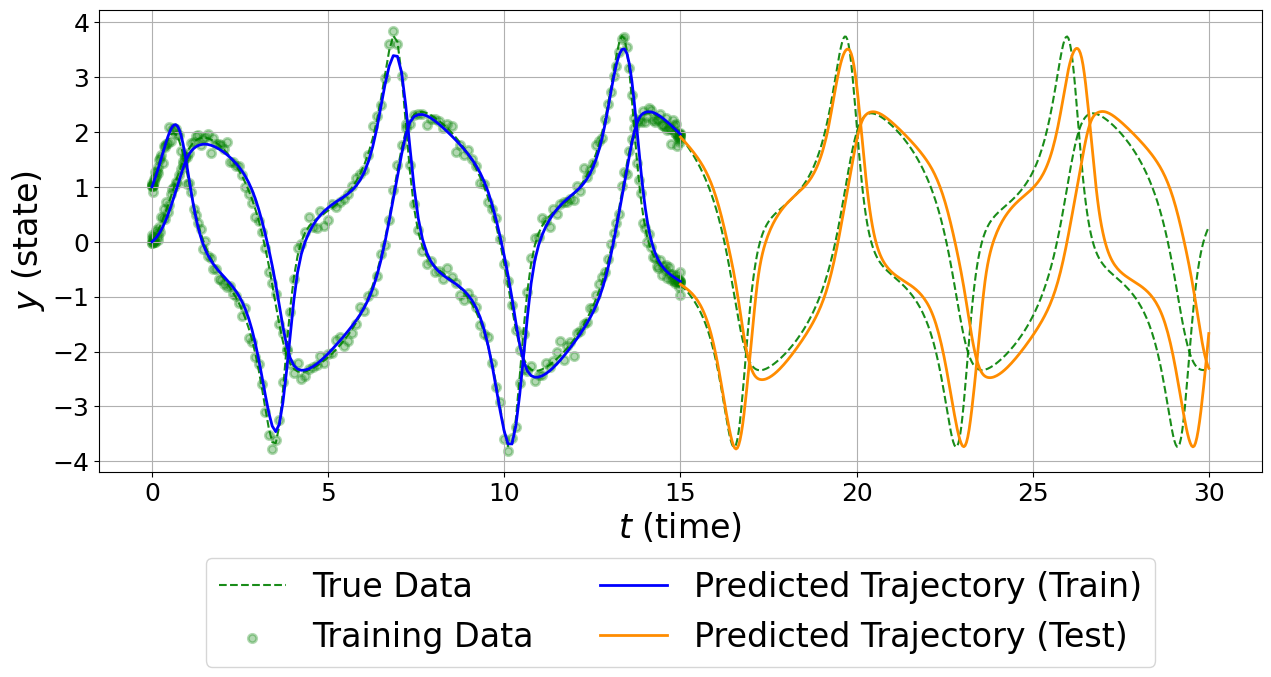}
    \caption{Predictions of a model trained with collocation-based method (Pyomo) for the \textbf{Van der Pol Oscillator}. 200 training points and 200 testing points.}
    \label{fig:pyomo_pred_vdp}
\end{figure}

After training with our proposed collocation-based framework, the learned Neural ODE is used in a standard ODE solver (JAX Diffrax) for forward simulation. 
Figure~\ref{fig:pyomo_pred_vdp} illustrates the prediction on both training and test ranges, showing that the collocation-trained Neural ODE captures the underlying dynamics effectively.

\subsubsection{Pre-Training Strategies}
While we found that collocation-based training generally converged to good solutions, the \textit{sequential} approach often resulted in premature local optima, perhaps owing to the feasible path approach. 
We therefore consider strategies for initial \textit{pre-training} on a subset of the data (20\%).
As shown in Table~\ref{tab:training_eval}, we use the notation \texttt{Model[Pre-training]} to denote different framework combinations, 
where \texttt{Model} is the strategy for the main training phase, and \texttt{Pre-training} is the strategy for the pre-training phase. 
The label \texttt{No} indicates that no pre-training was used.
The standard method for pre-training uses the same model for both phases ({JAX[JAX]} or {PyTorch[PyTorch]}).
We also explore a hybrid approach, where our Pyomo (collocation-based) model is used before continuing training with the benchmark models ({JAX[Pyomo]} or {PyTorch[Pyomo]}).
\begin{table}[h] 
\caption{Comparison of training frameworks. Smaller- and regular-size networks have layers of  $[2,8,2]$ and $[2,32,2]$ nodes, respectively.
Results averaged over 10 runs. Note the test MSEs are not available to the optimizers, which merely seek to minimize training MSE. } 
\label{tab:training_eval} 
\centering 
\begin{tabular}{@{}cccc@{}} \toprule 
\textbf{Model} [Pre-training] & \textbf{MSE Train} & \textbf{MSE Test} & \textbf{Time (s)} \\ \midrule 

\multicolumn{4}{l}{\textit{Shorter Training Duration: Regular-Size Network}} \\ \hdashline
\textbf{Pyomo} [No] & \textbf{0.0225} & \textbf{0.6139} & 7.144 \\ 
\textbf{JAX} [JAX] & 0.0553 & 1.0395 & 7.322 \\ \midrule
\multicolumn{2}{l}{\textit{Smaller-Size Network}} \\ \hdashline
\textbf{Pyomo} [No] & \textbf{0.0374} & \textbf{0.9792} &  2.043 \\ 
\textbf{JAX} [JAX] & 0.1003 & 1.2908 & 8.705 \\ \midrule

\multicolumn{4}{l}{\textit{Longer Training Duration: Regular-Size Network}} \\ \hdashline
\textbf{JAX} [No] & 2.7192 & 129.63 & 27.61 \\
\textbf{JAX} [JAX] & 0.0312 & 1.0954 & 23.85 \\
\textbf{JAX} [Pyomo] & \textbf{0.0098} & 0.4891 & 23.18 \\
\textbf{PyTorch} [PyTorch] & 0.4152 & 1.6728 & 25.14 \\ 
\textbf{PyTorch} [Pyomo] & 0.0111 & \textbf{0.4204} & 28.59 \\
\bottomrule 
\end{tabular}
\end{table}

\subsection{Performance Evaluation}

\subsubsection{Shorter Training Duration}
The collocation-based model exhibits fast convergence (approximately 7 seconds to train the regular-size network). 
We compare this with the accuracy achieved by the JAX model within the same training timeframe.
As demonstrated in the \textit{Shorter Training Duration} (top) section of Table \ref{tab:training_eval}, the Pyomo model achieves lower MSE during this time. 

\subsubsection{Smaller-Size Network}
We also observe that collocation-based training can produce models that achieve better performance with smaller network sizes, as shown in the \textit{Smaller-Size Network} (middle) section of Table \ref{tab:training_eval}. Despite letting the JAX model train for longer using the smaller-size network configuration, it does not reach the performance level of the Pyomo model. 
This suggests that the improved optimization framework results in better training of the limited model parameters, leading to lower training and testing MSEs. 

\subsubsection{Longer Training Duration}
\begin{figure}[h!]
    \centering
    \includegraphics[width=1\linewidth]
    {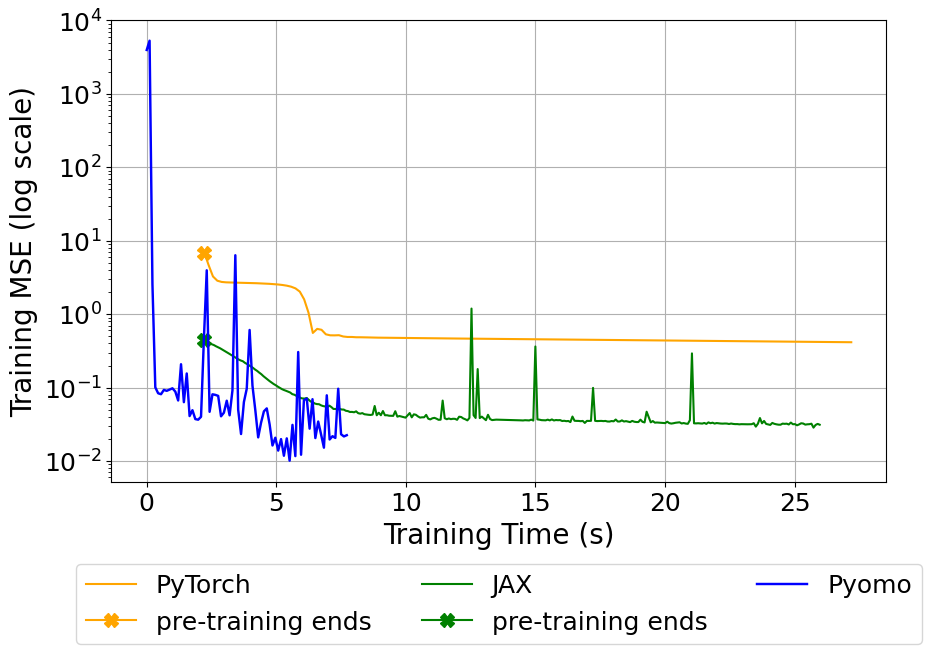}
    \caption{\textbf{Training} MSE for three training frameworks. Note that the MSE values at intermediate training times are obtained by interrupting IPOPT’s runtime and may appear unstable.}
    \label{fig:training_convergence_vdp}
\end{figure}

\begin{figure}[h!]
    \centering
    \includegraphics[width=1\linewidth]{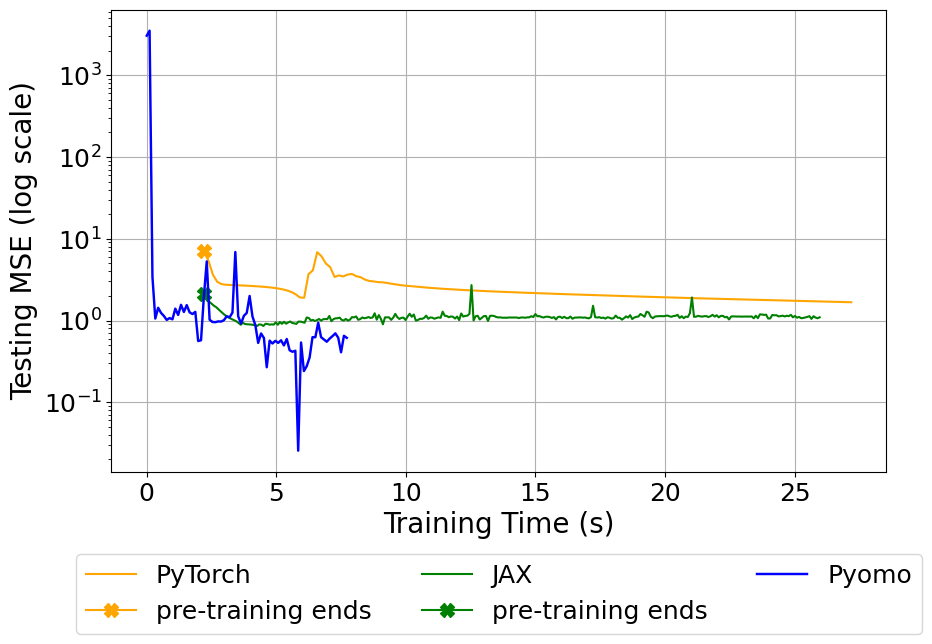}
    \caption{\textbf{Testing} MSE for three training frameworks.}
    \label{fig:testing_convergence_vdp}
\end{figure}

While training models with JAX and PyTorch for longer periods of time enhances their performance, we find they do not surpass the model trained with the collocation framework unless they are pre-trained with the latter, as detailed in \textit{Longer Training Duration} (bottom) section of Table \ref{tab:training_eval}. 
Figures \ref{fig:training_convergence_vdp} and \ref{fig:testing_convergence_vdp} show that the collocation-based training framework results in the fastest convergence in terms of both training and testing MSEs. 

\subsubsection{Hybrid Pre-Training}
\begin{figure}[h!]
    \centering
    \includegraphics[width=1\linewidth]{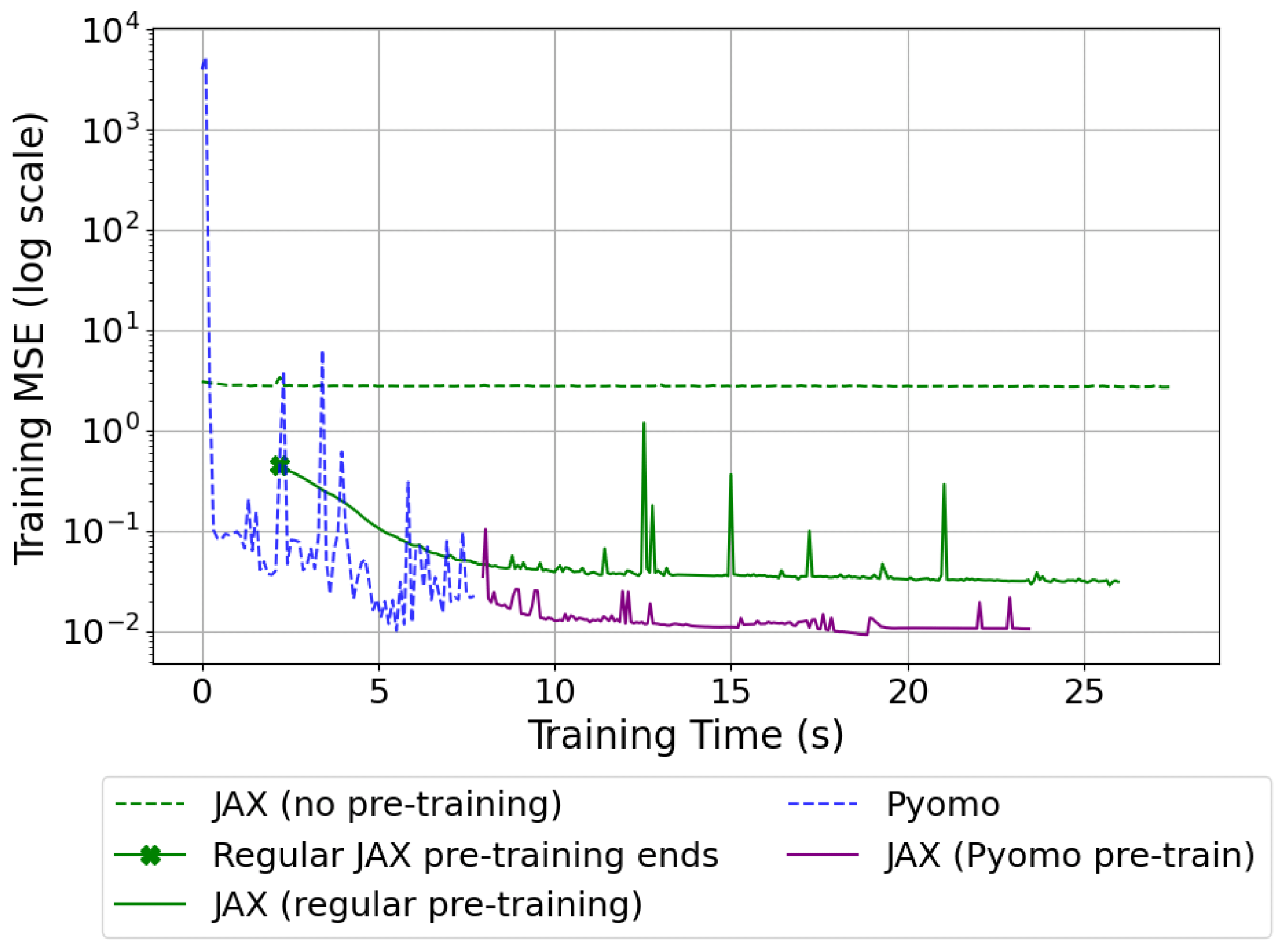}
    \caption{\textbf{Training} MSE of JAX model with pre-training.}
    \label{fig:pre_training}
\end{figure}

We also demonstrate that the proposed framework can be used to pre-train alongside existing methods. For example, after the collocation-based training converges, we use its weights and biases to initialize a JAX model. Subsequently, the JAX model can continue training and further improve the results, as seen in Figure \ref{fig:pre_training}. The combination of the collocation-based training followed by training using JAX achieves the lowest training MSE scores, as seen in Table \ref{tab:training_eval}. 

\subsubsection{Batching Using ADMM}
\begin{figure}[h!]
    \centering
    \includegraphics[width=0.80\linewidth]{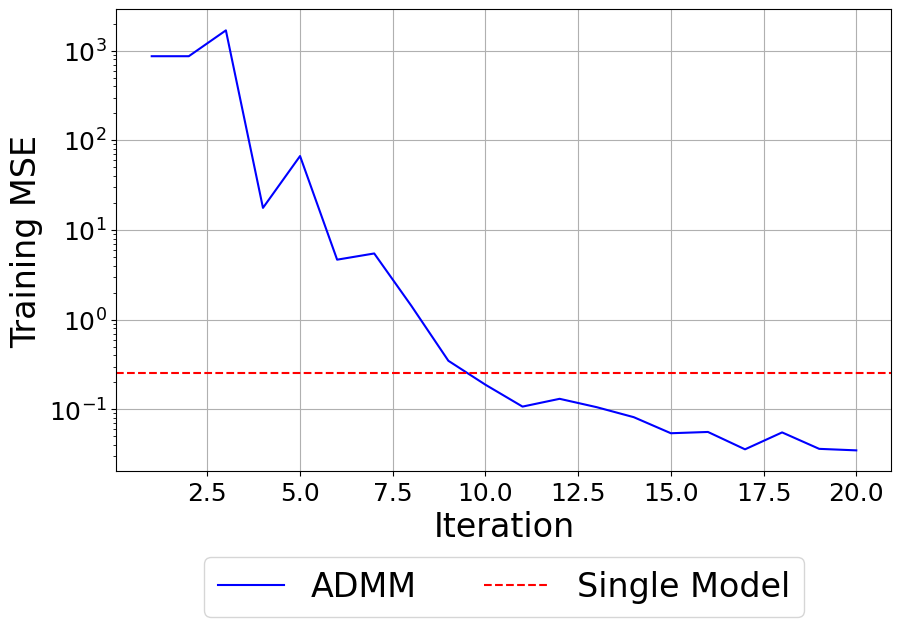}
    \caption{\textbf{ADMM} (150 + 150 training points) vs Single Model (300 training points) Performance}
    \label{fig:admm}
\end{figure}

Finally, we evaluate the performance of the proposed methodology for multi-batching using ADMM. Figure \ref{fig:admm} shows that ADMM is able to successfully coordinate the training of the two submodels, each containing half of the training data. The final MSE using the consensus weights of the ADMM-trained model also surpasses the performance of a monolithic-trained model. This improvement is perhaps related to folklore observations that stochastic (batched) gradients may help escape local optima. Ultimately, the ADMM framework provides an avenue to train larger models more effectively.    

\section{Conclusions}

In this work, we propose a collocation-based methodology using spectral methods for training Neural ODEs. By approximating the solution with Lagrange polynomials and enforcing differential equation constraints at collocation points, we recast the ODE problem as a system of algebraic constraints suitable for simultaneous optimization using IPOPT. Our experimental results on the Van der Pol Oscillator demonstrate that the proposed method achieves fast convergence and may enable more compact models compared to traditional sequential training approaches (e.g.,  implemented in JAX and PyTorch).

\bibliography{ifacconf}                         

\appendix
\end{document}